\documentclass[letterpaper]{article} 
\usepackage[preprint]{aaai2027}  
\usepackage[hyphens]{url}  
\usepackage{graphicx} 
\usepackage{natbib}  
\usepackage{caption} 
\usepackage{algorithm}
\usepackage{algorithmic}

\DeclareCaptionStyle{ruled}{labelfont=normalfont,labelsep=colon,strut=off} 

\usepackage{booktabs}

\usepackage{xcolor}    
\usepackage{colortbl}  
\usepackage{amsmath,amssymb}

\newcommand{\rank}{\operatorname{rank01}}
\definecolor{bestcolor}{HTML}{D1E7DD} 
\newcommand{\best}[1]{\cellcolor{bestcolor}\textbf{#1}}
\newcommand{\secondBest}[1]{\underline{#1}}

\title{RINSE: Robust Target-Time Normality Estimation\\for Zero-Shot Graph Anomaly Detection}

\author{
Taufikur Rahman Fuad\textsuperscript{\rm 1},
Md Abrar Jahin\textsuperscript{\rm 2},
Amir Hussain\textsuperscript{\rm 3}
}

\affiliations{
\textsuperscript{\rm 1}Islamic University of Technology\\
\textsuperscript{\rm 2}University of Southern California\\
\textsuperscript{\rm 3}King Fahd University of Petroleum \& Minerals\\
taufikur@iut-dhaka.edu, jahin@usc.edu, amir.hussain@kfupm.edu.sa
}

\begin{document}

\maketitle

\begin{abstract}
Zero-shot graph anomaly detection seeks to deploy a detector trained on source graphs to unseen, unlabeled targets, yet domain shift can make source-derived notions of normality unreliable. We introduce RINSE (Robust Iterative Normality Self-Estimation), a gradient-free target-time framework that keeps the source-trained detector fixed while sequentially estimating target normality, representation calibration, and evidence reliability from the target graph. Its core idea is to identify a reliable subset of low-residual target nodes, use them to construct a trimmed target-aware normality model, and combine complementary anomaly evidence through reliability-gated rank fusion and encoder ensembling. Across eight unseen target graphs, RINSE achieves the highest average AUPRC among the evaluated methods under two separate preprocessing protocols, while block ablations and sensitivity analyses support the combined design. These results support robust target-time estimation as a practical approach to generalist graph anomaly detection without target labels, gradients, or per-target tuning.
\end{abstract}

\section{Introduction}
Graph anomaly detection (GAD) identifies nodes in graphs whose attributes or connectivity deviate from normality. It supports fraud detection in financial and e-commerce networks, spam and bot detection on social platforms, and review manipulation detection~\citep{dominant2019,bwgnn2022,ghrn2023}. Because anomalies are less frequent and node-level labels are costly, the paradigm is often unsupervised, with a detector trained on each scored graph~\citep{qiaoDeepGraphAnomaly2025}. Reconstruction methods flag nodes reconstructed poorly by graph autoencoders~\citep{dominant2019,heADAGADAnomalyDenoisedAutoencoders2024}; contrastive methods detect nodes inconsistent with sampled contextual views~\citep{liu2022contrastive,slgad2021}; and affinity methods score disagreement with local neighborhoods~\citep{tam2023,care2025}.

Per-graph training incur structural costs. Each graph requires a separate training run, yet hyperparameters and stopping criteria cannot be validated without labels; label-free model selection remains an open problem~\citep{metaod2021,goswamiUnsupervisedModelSelection2023,ma2023need}. Unsupervised objectives are also optimized on contaminated data, allowing autoencoders to absorb the anomalies they aim to detect~\citep{heADAGADAnomalyDenoisedAutoencoders2024,kimRethinkingReconstructionbasedGraphLevel2024}. Generalist GAD reduces these costs by training one model on labeled source graphs and applying it to unseen targets~\citep{arc2024,panSurveyGeneralizationGraph2025}. In the zero-shot setting, the model ranks nodes in unseen target graphs without labels, fine-tuning, or target-specific configuration~\citep{unprompt2025,anomalygfm2025,zhangIAGGADZeroshotGeneralist2025,owleye2026}. This is the setting we study.

Zero-shot generalists encode anomaly criteria during source training as neighborhood prompts, graph-agnostic prototypes, or normal-pattern dictionaries, then apply them to unlabeled targets~\citep{unprompt2025,anomalygfm2025,owleye2026}. ARC uses labeled normal context at inference~\citep{arc2024}, while ARC$_{\mathrm{zero}}$ extends this direction by selecting pseudo-normal target context for label-free inference~\citep{arczero2026}. RINSE jointly estimates a trimmed target-normality dictionary, embedding calibration, and evidence reliability around a frozen detector.

Test-time adaptation offers a related route. Existing approaches optimize entropy, self-supervised objectives, graph transformations, or graph-level aligners using unlabeled test data~\citep{tent2021,ttt2020,gtrans2023,tune2026}. RINSE instead follows a gradient-free route: it estimates target normality from low-residual nodes and excludes high-residual candidates from subsequent calibration and dictionary updates. This connects zero-shot GAD with robust estimation under contaminated target data~\citep{diakonikolasRobustSparseEstimation2024,wangTractableNearOptimalAdversarial2023,diakonikolasEfficientMultivariateRobust2025}.

We ask whether a frozen zero-shot detector can estimate target normality, embedding scale, and evidence reliability from a contaminated, unlabeled target graph, without gradients or per-target tuning. We hypothesize that a robust, gradient-free estimator can estimate these quantities while limiting the influence of suspected anomalies. We present RINSE (\textbf{R}obust \textbf{I}terative \textbf{N}ormality \textbf{S}elf-\textbf{E}stimation), a test-time framework for a frozen truncated-attention detector. RINSE iteratively builds a target normality dictionary from low-residual nodes, calibrates embeddings to its statistics, and gates complementary evidence views by their agreement at the anchor ranking's extremes. Rank averaging across independently initialized encoders improves mean AUPRC under the evaluated protocol.

Our contributions are: \textbf{(i)} We cast zero-shot GAD around a frozen reconstruction scorer as contamination-controlled target-time normality estimation and develop a residual-trimmed iterative estimator without target labels or target-time gradients. \textbf{(ii)} We combine two-pass dictionary-statistics calibration with a label-free reliability gate over complementary reconstruction-based, distance-based, and affinity-based views, followed by rank-based encoder ensembling. \textbf{(iii)} We evaluate one shared configuration across eight unseen target graphs under both the standard and source-referenced preprocessing protocols, block ablations under both protocols, sensitivity analysis, and estimator diagnostics.

\section{Related Work}

\paragraph{Graph Anomaly Detection. }
Most deep GAD methods train a detector on the graph it scores~\citep{qiaoDeepGraphAnomaly2025}. Reconstruction methods flag poorly reconstructed nodes~\citep{dominant2019,heADAGADAnomalyDenoisedAutoencoders2024}; contrastive methods compare nodes with sampled contexts~\citep{liu2022contrastive,slgad2021}; affinity methods model neighborhood agreement~\citep{tam2023,care2025}; and supervised spectral methods capture the high-frequency patterns of anomalies~\citep{bwgnn2022,ghrn2023}. This target-specific paradigm requires label-free hyperparameter and stopping selection~\citep{metaod2021,goswamiUnsupervisedModelSelection2023,ma2023need} and learns from contaminated graphs, exposing reconstruction models to anomaly overfitting~\citep{heADAGADAnomalyDenoisedAutoencoders2024} and affinity models to violations of one-class homophily~\citep{tam2023,bwgnn2022,ghrn2023}. RINSE performs no target training: it treats contamination as a test-time estimation problem and limits suspected anomalies' influence on normality estimation.

\paragraph{Generalist and Zero-Shot Cross-Domain GAD. }
Earlier cross-domain GAD methods adapt source knowledge to a given target through adversarial~\citep{commander2022} or anomaly-aware contrastive alignment~\citep{act2023}, with later transfer-based variants~\citep{huangCrossDomainGraphAnomaly2025}. They access the target graph during training and learn a separate model for each target. Broader graph-transfer methods use language-encoded attributes~\citep{liZeroGInvestigatingCrossdataset2024,wangUniGTEUnifiedGraphText2025} or few-shot prompt learning~\citep{yuHGPromptBridgingHomogeneous2024,fuGraphPromptingGraph2025}, and thus address different input or supervision settings.

ARC established the generalist GAD setting by training once on labeled source graphs and scoring unseen targets without retraining, using a few labeled normal nodes at inference~\citep{arc2024}. ARC$_{\mathrm{zero}}$ later enables label-free inference by selecting representative pseudo-normal target context~\citep{arczero2026}. Other zero-shot methods learn neighborhood prompts~\citep{unprompt2025}, align graph-agnostic prototypes with residual features~\citep{anomalygfm2025}, combine invariant representations with affinity learning~\citep{zhangIAGGADZeroshotGeneralist2025}, or reconstruct target nodes through truncated attention over a source-trained normality dictionary~\citep{owleye2026}. RINSE differs by jointly estimating target normality, embedding calibration, and evidence reliability through a gradient-free procedure around a fixed source-trained scorer.

\paragraph{Test-Time Adaptation and Unsupervised Model Selection. }
Test-time adaptation updates deployed systems using unlabeled test data. Test-time training optimizes self-supervised losses~\citep{ttt2020,liuTTTWhenDoes2021}, Tent minimizes prediction entropy through normalization parameters~\citep{tent2021}, and GTrans adapts target features and graph structure~\citep{gtrans2023}. TUNE adapts pretrained graph anomaly detectors to unseen normal patterns by optimizing a graph-level aligner~\citep{tune2026}. Recent backpropagation-free approaches align target subspaces or feature statistics~\citep{wangBackpropagationfreeNetwork3D2024,dongEBATSEfficientBackpropagationFree2025a,zhangBackpropagationFreeTestTimeAdaptation2025a} or select informative tokens~\citep{yazdanpanahPurgeGateBackpropagationFreeTestTime2025a}. RINSE requires no target-time optimization and instead performs forward-only estimation for contaminated one-class ranking.

Label-free model selection uses meta-learning~\citep{metaod2021}, surrogate metrics~\citep{goswamiUnsupervisedModelSelection2023}, or internal evaluation criteria~\citep{ma2023need}, while uncertainty-aware view rejection has been studied for supervised multi-view classification~\citep{liuEnhancingMultiviewClassification2025}. RINSE separates these roles: held-out source validation selects the checkpoint, while the unlabeled target graph determines view reliability at inference.

\begin{figure*}[t]
\centering
\includegraphics[width=\textwidth]{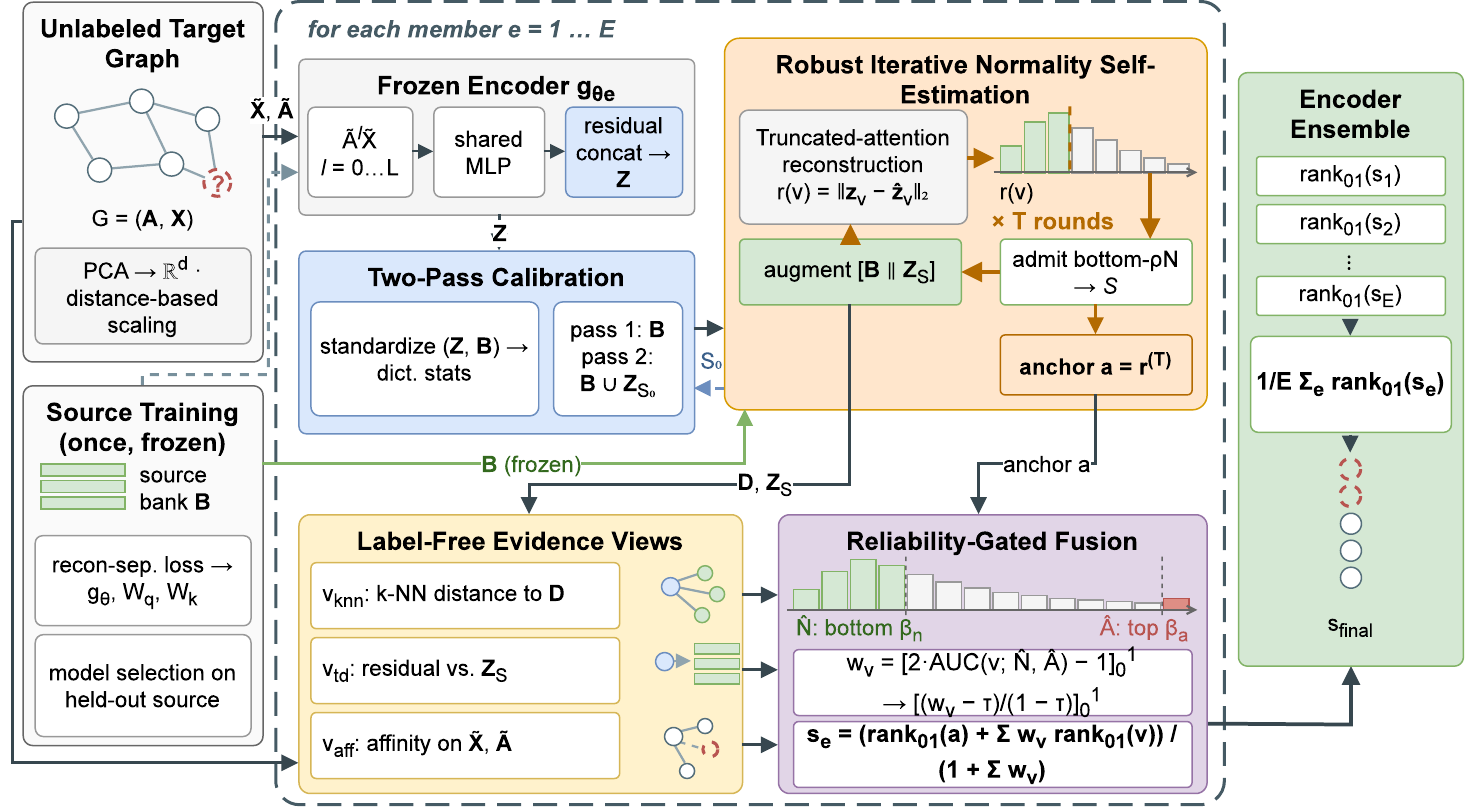}
\caption{Overview of RINSE. For each ensemble member, the encoder $g_{\theta_e}$, attention projections $\mathbf{W}_q,\mathbf{W}_k$, and source bank $\mathbf{B}_e$ are learned or constructed from source graphs and then frozen. All target-time components operate independently on the unlabeled target graph, without labels, gradients, or per-target tuning. The blue dashed path supplies provisional normals $\mathcal{S}_0$ to the second calibration pass; red dashed nodes denote suspected anomalies ranked highest at output.}
\label{fig:overview}
\end{figure*}

\section{Preliminaries}
\label{sec:prelim}
Let $G=(\mathcal{V},\mathcal{E},\mathbf{X})$ be an attributed graph with $N=|\mathcal{V}|$ nodes, adjacency matrix $\mathbf{A}\in\{0,1\}^{N\times N}$, and attributes $\mathbf{X}\in\mathbb{R}^{N\times d_G}$. Each node has a latent label $y_v\in\{0,1\}$ ($1=$ anomalous), with $\sum_v y_v\ll N$. In \emph{zero-shot generalist graph anomaly detection}~\citep{arc2024,unprompt2025}, a model is trained once on labeled source graphs $\mathcal{D}_{\mathrm{src}}=\{G^{(1)},\dots,G^{(M)}\}$ and applied to unlabeled target graphs from unseen domains. For each target, it produces scores $s:\mathcal{V}\rightarrow\mathbb{R}$ that rank anomalous nodes above normal ones, without target labels, fine-tuning, or target-specific configuration. Because feature spaces differ across domains, each graph is projected to a shared dimension $d$ using PCA, preceded by Gaussian random projection when $d_G<d$, and rescaled by distance-based normalization following~\citep{arc2024,owleye2026}. We denote the processed attributes and normalized adjacency by $\tilde{\mathbf{X}}$ and $\tilde{\mathbf{A}}$.

\section{Methodology}

RINSE performs robust target-time estimation around a frozen detector. For each ensemble member $e$, it encodes the target graph with $g_{\theta_e}$, calibrates target and source-bank embeddings in two passes using provisional low-residual normals, iteratively augments the source bank with trimmed target normals to obtain an anchor score and admitted set, computes three complementary evidence views, and fuses their rank-normalized scores with label-free reliability weights. The final prediction averages score ranks across $E$ independently initialized members. All target-time steps use only the unlabeled target graph, require no gradients, and share one configuration across targets. Figure~\ref{fig:overview} shows the pipeline, and Algorithm~\ref{alg:rinse} summarizes inference.

\subsection{Frozen Truncated-Attention Scorer}
RINSE uses a source-trained scorer without target-time updates. A shared MLP $g_\theta$ encodes propagated attributes,
$\mathbf{Z}^{[l]}=g_\theta(\tilde{\mathbf{A}}^l\tilde{\mathbf{X}})$, and forms
\begin{equation}
\mathbf{Z}=
\big[\mathbf{Z}^{[1]}-\mathbf{Z}^{[0]}\;\|\;\cdots\;\|\;
\mathbf{Z}^{[L]}-\mathbf{Z}^{[0]}\big]
\in\mathbb{R}^{N\times Lh}.
\end{equation}
Given a normality dictionary
$\mathbf{P}=[\mathbf{p}_1;\dots;\mathbf{p}_m]$, let
$\mathbf{q}_v=\sigma(\mathbf{W}_q\mathbf{z}_v)$,
$\mathbf{k}_j=\sigma(\mathbf{W}_k\mathbf{p}_j)$, and
$e_{vj}=\langle\mathbf{q}_v,\mathbf{k}_j\rangle/\sqrt{Lh}$,
where $\sigma$ is LeakyReLU. After masking the smallest
$\lfloor\eta m\rfloor$ logits,
\begin{equation}
\hat{\mathbf{z}}_v
=\sum\nolimits_j\alpha_{vj}\mathbf{p}_j,
\qquad
\alpha_{vj}=\operatorname{softmax}_j(e_{vj}/\gamma),
\end{equation}
and the reconstruction residual is
\begin{equation}
r(v;\mathbf{P})
=\lVert\mathbf{z}_v-\hat{\mathbf{z}}_v\rVert_2.
\label{eq:residual}
\end{equation}
A query's own dictionary entry is masked when present. RINSE freezes this scorer and estimates its target-time dictionary, calibration, and evidence weights.

\begin{algorithm}[t]
\caption{RINSE inference. \textsc{Estimate} implements Eq.~\ref{eq:iterate}; \textsc{Calibrate} applies Eq.~\ref{eq:calib} to target and bank embeddings using reference $\mathbf{C}$; $\rank(\cdot)$ maps scores to $[0,1]$, and $[\cdot]_0^1$ clips to this interval.}
\label{alg:rinse}
\textbf{Input}: target graph $G$; frozen members
$\{\mathcal{M}_e\}_{e=1}^{E}$ with source banks $\{\mathbf{B}_e\}$;
parameters $\rho,T,k,R,\beta_n,\beta_a,\tau,\varepsilon$\\
\textbf{Output}: $s_{\mathrm{final}}\in[0,1]^N$
\begin{algorithmic}[1]
\FOR{$e=1$ \TO $E$}
    \STATE $\mathbf{Z}\leftarrow g_{\theta_e}(G)$;\quad
    $\mathbf{B}\leftarrow\mathbf{B}_e$
    \STATE $(\mathbf{Z}',\mathbf{B}')
    \leftarrow\textsc{Calibrate}(\mathbf{Z},\mathbf{B};\mathbf{B},\varepsilon)$;\quad
    $(\_,\mathcal{S}_0)\leftarrow
    \textsc{Estimate}(\mathbf{Z}',\mathbf{B}';\rho,T)$
    \STATE $(\mathbf{Z},\mathbf{B})
    \leftarrow\textsc{Calibrate}\!\left(
    \mathbf{Z},\mathbf{B};
    [\mathbf{B}\,\|\,\mathbf{Z}_{\mathcal{S}_0}],\varepsilon
    \right)$
    \STATE $(a,\mathcal{S})
    \leftarrow\textsc{Estimate}(\mathbf{Z},\mathbf{B};\rho,T)$;\quad
    $\mathbf{D}\leftarrow[\mathbf{B}\,\|\,\mathbf{Z}_{\mathcal{S}}]$
    \STATE $\hat{\mathcal{N}}
    \leftarrow\{u:\rank(a)(u)<\beta_n\}$;\quad
    $\hat{\mathcal{A}}
    \leftarrow\{u:\rank(a)(u)\ge
    1-\max(\beta_a,10/N)\}$
    \STATE compute
    $\mathcal{V}=\{v_{\mathrm{knn}},v_{\mathrm{td}},v_{\mathrm{aff}}\}$
    from $\mathbf{D},\mathbf{Z}_{\mathcal{S}},G$ using $k,R$
    \FORALL{$v\in\mathcal{V}$}
        \STATE $w_v\leftarrow
        \left[
        \dfrac{
        [\,2\,\mathrm{AUC}(v;\hat{\mathcal{N}},\hat{\mathcal{A}})-1\,]_0^1-\tau
        }{1-\tau}
        \right]_0^1$
    \ENDFOR
    \STATE $s_e\leftarrow
    \dfrac{\rank(a)+\sum_{v\in\mathcal{V}}w_v\rank(v)}
    {1+\sum_{v\in\mathcal{V}}w_v}$
\ENDFOR
\STATE \textbf{return}\quad
$s_{\mathrm{final}}
=\dfrac{1}{E}\sum_{e=1}^{E}\rank(s_e)$
\end{algorithmic}
\end{algorithm}

\subsection{Robust Iterative Normality Self-Estimation}
\label{sec:estimator}
For one ensemble member, let $\mathbf{B}$ be its frozen source bank, with the member index suppressed. To adapt source normality to the target while limiting anomaly influence, RINSE repeatedly admits low-residual target nodes. Starting from $r^{(0)}(v)=r(v;\mathbf{B})$, for $t=1,\dots,T$ it computes
\begin{equation}
\begin{aligned}
\mathcal{S}^{(t)}
&=\operatorname*{arg\,bottom}_{\rho N} r^{(t-1)}\\
r^{(t)}(v)
&=r\!\left(v;\,[\mathbf{B}\,\|\,\mathbf{Z}_{\mathcal{S}^{(t)}}]\right)
\end{aligned}
\label{eq:iterate}
\end{equation}
Each round combines the $\rho N$ lowest-residual target embeddings with $\mathbf{B}$ and rescores all nodes with self-masking. The final anchor is $a(v)=r^{(T)}(v)$; its admitted set
$\mathcal{S}=\operatorname*{arg\,bottom}_{\rho N}a$
defines the target-aware dictionary
$\mathbf{D}=[\mathbf{B}\,\|\,\mathbf{Z}_{\mathcal{S}}]$.
Thus, $\rho$ trims the highest-residual $1-\rho$ fraction from each update. The admitted sets remain approximately $97$--$99\%$ normal across targets and rounds.

\subsection{Two-Pass Dictionary-Statistics Calibration}
\label{sec:calib}
Embedding-scale shift affects attention, distance, and residual scores. For reference matrix $\mathbf{C}$, let $\boldsymbol{\mu}_{\mathbf C}$ and $\boldsymbol{\sigma}_{\mathbf C}$ be its coordinate-wise moments, and define
\begin{equation}
\operatorname{Cal}_{\mathbf C}(\mathbf{X})
=
c\,(\mathbf{X}-\boldsymbol{\mu}_{\mathbf C})
\oslash
\big(\boldsymbol{\sigma}_{\mathbf C}
+\varepsilon\bar{\sigma}_{\mathbf C}\big),
\label{eq:calib}
\end{equation}
where $\bar{\sigma}_{\mathbf C}$ is the mean coordinate standard deviation and $c$ restores the source bank's mean row norm. The same transformation is applied to $\mathbf{Z}$ and $\mathbf{B}$.

Calibration uses two passes. First, $\mathbf{C}=\mathbf{B}$ yields provisional normals $\mathcal{S}_0$ through Section~\ref{sec:estimator}. The original embeddings are then recalibrated with
$\mathbf{C}=[\mathbf{B}\,\|\,\mathbf{Z}_{\mathcal{S}_0}]$,
allowing low-residual target patterns to refine the statistics without exposing them to high-residual nodes.

\subsection{Label-Free Multi-View Evidence}
\label{sec:views}
The anchor reflects attention-reconstruction geometry; RINSE adds three complementary label-free views.

\textbf{$k$NN distance. }
The mean distance to the $k$ nearest patterns in the target-aware dictionary is
\begin{equation}
v_{\mathrm{knn}}(u)
=
\frac{1}{k}
\sum_{\mathbf{p}\in\mathcal{N}_k(\mathbf{z}_u;\mathbf{D})}
\lVert\mathbf{z}_u-\mathbf{p}\rVert_2,
\end{equation}
excluding the query's own entry when present. This view measures proximity to individual normal patterns rather than their attention-weighted combination.

\textbf{Target-dictionary residual. }
The target-only score
$v_{\mathrm{td}}(u)=r(u;\mathbf{Z}_{\mathcal{S}})$
reconstructs each node from admitted target normals, with self-masking, and removes dependence on the source bank.

\textbf{Raw-attribute affinity. }
Let $\bar{\mathbf{X}}$ be the row-normalized processed attributes. Initialize
$v_{\mathrm{aff}}^{(0)}(u)
=1-\cos\!\big(\bar{\mathbf{x}}_u,
(\tilde{\mathbf{A}}\bar{\mathbf{X}})_u\big)$.
For $q=0,\dots,R-1$, set
$\boldsymbol{\omega}^{(q)}
=1-\rank(v_{\mathrm{aff}}^{(q)})$ and update
\begin{equation}
v_{\mathrm{aff}}^{(q+1)}(u)
=
1-\cos\!\left(
\bar{\mathbf{x}}_u,
\frac{
\big(\tilde{\mathbf{A}}
(\boldsymbol{\omega}^{(q)}\odot\bar{\mathbf{X}})\big)_u
}{
(\tilde{\mathbf{A}}\boldsymbol{\omega}^{(q)})_u
}
\right).
\end{equation}
The final view is $v_{\mathrm{aff}}=v_{\mathrm{aff}}^{(R)}$. Rank weighting reduces the influence of high-scoring neighbors, while operating outside the encoder provides complementary attribute-structural evidence.

\begin{table*}[t]
\centering
\small
\setlength{\tabcolsep}{1pt}
\begin{tabular*}{\textwidth}{@{\extracolsep{\fill}}l*{8}{r}r}
\toprule
Method      & Cora         & Flickr      & ACM      & BlogCatalog     & Facebook       & Weibo     & Reddit     & Amazon  & Average \\
\midrule
\multicolumn{10}{c}{\textit{Supervised (10-shot)}} \\
BWGNN  &  9.57\(\pm\)2.40   & 12.39\(\pm\)2.68   & 13.37\(\pm\)6.03 & 12.97\(\pm\)3.15    &  5.81\(\pm\)1.17    &  9.55\(\pm\)2.12    &  3.21\(\pm\)2.32    &  12.40\(\pm\)1.86    &  9.91    \\ 
GHRN  &  14.04\(\pm\)0.73    &  16.45\(\pm\)2.59    &  16.29\(\pm\)1.41    &  13.58\(\pm\)2.19    &  6.24\(\pm\)1.12    &  17.51\(\pm\)1.52    &  \secondBest{4.44\(\pm\)1.15}    &  13.84\(\pm\)2.63    &  12.80    \\
\midrule
\multicolumn{10}{c}{\textit{Unsupervised}} \\
DOMINANT  & 31.77\(\pm\)0.34  & {28.76\(\pm\)1.52}  & 32.49\(\pm\)4.97  & {29.51\(\pm\)3.44}  & 3.42\(\pm\)0.86  & 29.63\(\pm\)0.86  & 3.28\(\pm\)0.37 &  36.80\(\pm\)8.37 &  24.46    \\
SL-GAD  & 18.27\(\pm\)1.01  & 16.93\(\pm\)8.20  & 1.33\(\pm\)0.23  & 9.47\(\pm\)3.00  & 0.93\(\pm\)0.23  & 35.80\(\pm\)1.41  & 4.00\(\pm\)2.27 &  5.33\(\pm\)2.20 &  11.51    \\ 
TAM  & 9.43\(\pm\)0.27  & 23.34\(\pm\)1.42  & \secondBest{40.68\(\pm\)2.58}  & 25.59\(\pm\)4.76  & \best{12.18\(\pm\)3.14}  & 23.01\(\pm\)15.14  & 4.22\(\pm\)0.22 & {45.26\(\pm\)4.34} &  22.96    \\ 
CARE  & {35.12\(\pm\)0.23}  & 25.64\(\pm\)0.16  & 37.76\(\pm\)0.35  & 25.06\(\pm\)0.10  & 5.52\(\pm\)0.34  & {40.70\(\pm\)0.74}  & 3.17\(\pm\)0.17 &  56.76\(\pm\)1.44 &  28.72    \\
\midrule
\multicolumn{10}{c}{\textit{One-for-all}} \\
ARC & \secondBest{45.20\(\pm\)1.08}  & 35.13\(\pm\)0.20  & {39.02\(\pm\)0.08}  & 33.43\(\pm\)0.15  & 4.25\(\pm\)0.47  & \best{64.18\(\pm\)0.68}  & 4.20\(\pm\)0.25 &  20.48\(\pm\)6.89 &  30.74    \\ 
UNPrompt  & 9.84\(\pm\)2.90  & 25.21\(\pm\)1.84  & 11.18\(\pm\)1.67  & 18.24\(\pm\)13.05  & 4.32\(\pm\)0.55  & 20.58\(\pm\)5.62  & 3.77\(\pm\)0.32 &  9.41\(\pm\)2.69 &  12.82    \\
OWLEYE  & 43.94\(\pm\)0.46  & \secondBest{37.69\(\pm\)0.25}  & 39.75\(\pm\)0.13  & \secondBest{34.99\(\pm\)0.31}  & 5.62\(\pm\)1.17  & 60.90\(\pm\)0.21  & 4.25\(\pm\)0.11 &  \best{62.20\(\pm\)3.18} &  \secondBest{36.17}    \\
\midrule
\textbf{RINSE}  & \best{50.44\(\pm\)0.45}  & \best{38.64\(\pm\)0.12}  & \best{48.32\(\pm\)0.04}  & \best{38.12\(\pm\)0.16}  & \secondBest{9.89\(\pm\)1.34}  & \secondBest{63.69\(\pm\)0.23}  & \best{4.80\(\pm\)0.16} &  \secondBest{59.57\(\pm\)7.38} &  \best{39.18}    \\
\bottomrule
\end{tabular*}
\caption{AUPRC (\%) under the standard normalization protocol, reported as mean\(\pm\)standard~deviation. The final column averages the eight target graphs. Best results are highlighted in bold and second-best results are underlined.}
\label{tab:standard_auprc}
\end{table*}

\begin{table*}[t]
\centering
\small
\setlength{\tabcolsep}{1pt}
\begin{tabular*}{\textwidth}{@{\extracolsep{\fill}}l*{8}{r}r}
\toprule
Method      & Cora         & Flickr      & ACM      & BlogCatalog     & Facebook       & Weibo     & Reddit     & Amazon  & Average \\
\midrule
ARC       & \secondBest{44.90\(\pm\)0.51} & \secondBest{38.04\(\pm\)0.26} & \secondBest{40.80\(\pm\)0.12} & \secondBest{35.71\(\pm\)0.22} & \secondBest{6.95\(\pm\)2.03} & \secondBest{62.39\(\pm\)0.45} & 4.04\(\pm\)0.18 & 22.54\(\pm\)7.21 & 31.92 \\
UNPrompt  & 7.32\(\pm\)0.85 & 22.42\(\pm\)1.45 & 12.98\(\pm\)0.80 & 28.93\(\pm\)0.65 & 6.24\(\pm\)1.50 & 23.67\(\pm\)2.21 & \secondBest{4.79\(\pm\)0.29} & 28.19\(\pm\)11.77 & 16.82 \\
OWLEYE    & 44.18\(\pm\)1.20 & 37.77\(\pm\)0.49 & 39.87\(\pm\)0.17 & 34.88\(\pm\)0.20 & 4.57\(\pm\)0.56 & 61.15\(\pm\)0.56 & 4.10\(\pm\)0.09 & \secondBest{59.83\(\pm\)1.24} & \secondBest{35.79} \\
\midrule
\textbf{RINSE}   & \best{49.87\(\pm\)1.72} & \best{38.62\(\pm\)0.05} & \best{48.28\(\pm\)0.30} & \best{38.29\(\pm\)0.37} & \best{9.21\(\pm\)2.29} & \best{63.23\(\pm\)0.34} & \best{5.04\(\pm\)0.31} & \best{63.89\(\pm\)2.48} & \best{39.55} \\
\bottomrule
\end{tabular*}
\caption{AUPRC (\%) under the leak-free normalization protocol, reported as mean\(\pm\)standard~deviation. The final column averages the eight target graphs. Best results are highlighted in bold and second-best results are underlined.}
\label{tab:leakfree_auprc}
\end{table*}

\subsection{Reliability-Gated Fusion and Encoder Ensembling}
\label{sec:gate}
For each ensemble member, RINSE estimates view reliability from agreement with the anchor at its ranking extremes. Suppressing the member index, define
\[
\begin{aligned}
\hat{\mathcal{N}}
&=\{u:\rank(a)(u)<\beta_n\}\\
\hat{\mathcal{A}}
&=\{u:\rank(a)(u)\ge 1-\max(\beta_a,10/N)\}
\end{aligned}
\]
For each
$v\in\mathcal{V}
=\{v_{\mathrm{knn}},v_{\mathrm{td}},v_{\mathrm{aff}}\}$,
\begin{equation}
w_v
=
\left[
\frac{
\big[2\,\mathrm{AUC}(v;\hat{\mathcal{N}},\hat{\mathcal{A}})-1\big]_0^1-\tau
}{
1-\tau
}
\right]_0^1,
\label{eq:pauc}
\end{equation}
where $[\cdot]_0^1$ clips to $[0,1]$. The resulting per-member score is
\begin{equation}
s(u)
=
\frac{
\rank(a)(u)+\sum_{v\in\mathcal{V}}w_v\rank(v)(u)
}{
1+\sum_{v\in\mathcal{V}}w_v
}.
\label{eq:fuse}
\end{equation}
The anchor retains unit weight, so unreliable views can be suppressed without replacing the primary score.

RINSE runs this pipeline independently for $E$ members and averages their normalized ranks:
\begin{equation}
s_{\mathrm{final}}(u)
=
\frac{1}{E}\sum_{e=1}^{E}\rank(s_e)(u).
\end{equation}
Rank normalization makes member scores comparable before averaging.

\begin{figure*}[t]
  \centering
  \includegraphics[width=\textwidth]{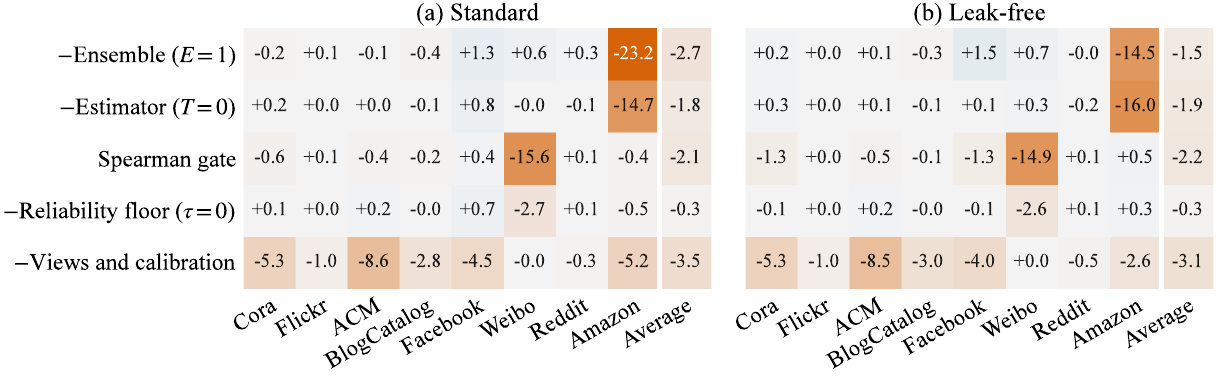}
\caption{Component ablation under the (a) standard and (b) leak-free protocols. Each cell reports the AUPRC change relative to full RINSE when a component is removed or replaced; the final column averages the eight target graphs. Both panels use the same scale.}
\label{fig:ablation}
\end{figure*}

\subsection{Source Training and Model Selection}
\label{sec:training}
We train $g_\theta,\mathbf{W}_q,\mathbf{W}_k$ on labeled source graphs using the reconstruction-separation objective of~\citet{owleye2026}. Sampled normal embeddings form the reconstruction dictionary. For normal and anomalous queries $\mathbf{z}_n,\mathbf{z}_a$, the loss is
\begin{equation}
\begin{aligned}
\mathcal{L}={}&
1-\cos(\mathbf{z}_n,\hat{\mathbf{z}}_n)
+\big[\cos(\mathbf{z}_a,\hat{\mathbf{z}}_a)\big]_+\\
&+\big[
\lVert\hat{\mathbf{z}}_n-\mathbf{z}_n\rVert_2
-\lVert\hat{\mathbf{z}}_n-\mathbf{z}_a\rVert_2
+\delta
\big]_+ ,
\end{aligned}
\end{equation}
where $\delta$ is the separation margin.

Checkpoint selection uses a fixed held-out source domain. For each ensemble member, we train on PubMed, Questions, and YelpChi, evaluate the full single-member inference pipeline on CiteSeer after each epoch, and retain the checkpoint with the highest AUPRC. We then freeze the scorer and rebuild its source bank from all four source graphs. No target graph enters training, checkpoint selection, or bank construction, and all targets share one configuration.

\section{Experiments}
\label{sec:experiments}

\subsection{Experimental Setup}
\paragraph{Datasets. }We follow the experimental and evaluation protocol of \citet{owleye2026}. RINSE and the one-for-all baselines use four source graphs and are evaluated on eight disjoint target graphs. The source set $\mathcal{T}_{\mathrm{train}}=\{\text{PubMed, CiteSeer, Questions, YelpChi}\}$ covers citation, question-answering, and co-review networks with injected or organic anomalies. The target set $\mathcal{T}_{\mathrm{test}}=
\{\text{Cora},\allowbreak\ \text{Flickr},\allowbreak\ \text{ACM},
\allowbreak\ \text{BlogCatalog},\allowbreak\ \text{Facebook},
\allowbreak\ \text{Weibo},\allowbreak\ \text{Reddit},
\allowbreak\ \text{Amazon}\}$ contains graphs not used for training or model selection.

\paragraph{Baselines. }We compare RINSE with supervised, unsupervised, and one-for-all GAD methods. The supervised baselines, BWGNN~\citep{bwgnn2022} and GHRN~\citep{ghrn2023}, use the 10-shot protocol, with five labeled normal and five labeled anomalous nodes per target graph. The unsupervised baselines are DOMINANT~\citep{dominant2019}, SL-GAD~\citep{slgad2021}, TAM~\citep{tam2023}, and CARE~\citep{care2025}; the one-for-all baselines are ARC~\citep{arc2024}, UNPrompt~\citep{unprompt2025}, and OWLEYE~\citep{owleye2026}.

\begin{figure*}[t]
  \centering
  \includegraphics[width=\textwidth]{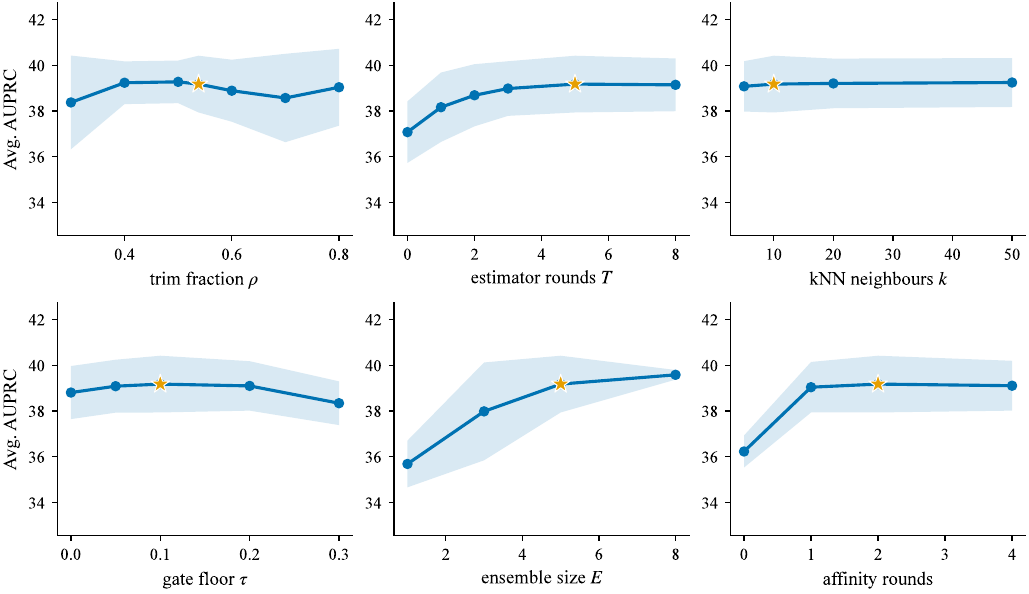}
\caption{Sensitivity of RINSE to six inference settings. Each panel varies one setting while fixing the others at their defaults, marked by an orange star. Curves report mean AUPRC across the eight target graphs, with shading for one standard deviation over five independent runs.}  \label{fig:sensitivity}
\end{figure*}

\paragraph{Implementation Details. }
Each ensemble member trains $g_\theta,\mathbf{W}_q,\mathbf{W}_k$ on PubMed, Questions, and YelpChi, using CiteSeer for checkpoint selection as described in Section~\ref{sec:training}. After restoring the best checkpoint from $10$ epochs, its source bank is rebuilt from all four source graphs. Target graphs provide neither training nor selection labels. All graphs use processed attribute dimension $d=64$. The encoder is a two-layer MLP over $L=4$ propagation hops, with hidden width $384$, ELU activation, and dropout $0.1$. Truncated attention uses a bank of $2048$ labeled-normal embeddings, truncation ratio $\eta=0.7$, temperature $\gamma=0.031$, and up to $1024$ normal dictionary embeddings per source graph during training.

At inference, RINSE uses $\rho=0.54$, $T=5$, $k=10$, and $R=2$ affinity refinements. The reliability gate uses $\beta_n=0.5$, $\beta_a=0.05$, and $\tau=0.1$; calibration uses $\varepsilon=0.05$. We rank-average $E=5$ independently initialized members. Each member is optimized with Adam using learning rate $1.14\times10^{-3}$, weight decay $9.4\times10^{-5}$, and source-loss margin $\delta=0.2$. Hyperparameters were selected using held-out source validation, without inspecting target labels or target performance, and the resulting shared configuration was fixed for all target graphs. We report five independent runs; member $e$ in run $t$ uses seed $100t+e$.  Source training is repeated for each member, while the complete target-time pipeline is gradient-free. We report AUROC and AUPRC, using AUPRC as the primary metric because anomalies are rare and it emphasizes performance on the minority class.

\subsection{Results}

\paragraph{Main Results. }
Table~\ref{tab:standard_auprc} shows that RINSE achieves the highest mean average precision among the evaluated methods, reaching $39.18$ and improving over the matched one-for-all baseline OWLEYE by $3.01$ points. RINSE ranks first on five targets and second on the remaining three. It increases performance over OWLEYE on seven of eight targets, with a median paired gain of $2.96$ points, and attains the best mean rank among the ten evaluated methods ($1.38$ versus $2.75$ for OWLEYE). Treating the eight target datasets as blocks, a Friedman test indicates significant differences across the evaluated methods ($p=3.6\times10^{-6}$). In the matched comparison with OWLEYE, an exact two-sided Wilcoxon signed-rank test gives $p=0.039$. 

\paragraph{Leak-Free Evaluation. }
We use \emph{leak-free} as shorthand for the source-referenced normalization path, in which cross-graph reference statistics are computed only from the source graphs and each target is preprocessed independently of the other target graphs. Under this protocol, RINSE ranks first among the four evaluated one-for-all methods on all eight targets and achieves a mean average precision of $39.55$ (Table~\ref{tab:leakfree_auprc}). Its margin over OWLEYE increases from $3.01$ points under standard normalization to $3.76$ points, showing that the matched-protocol improvement persists under source-referenced preprocessing. Because this preprocessing comparison is defined for the one-for-all pipeline, supervised and conventional unsupervised baselines are not repeated in Table~\ref{tab:leakfree_auprc}.

\paragraph{Ablation Study. }
Figure~\ref{fig:ablation} evaluates the contribution of each target-time component under both normalization protocols. The evidence views and dictionary-statistics calibration provide the largest average gain, improving AUPRC by $3.5$ and $3.1$ points under the standard and leak-free protocols, respectively, with particularly strong contributions on Cora, ACM, and Facebook. Ensembling and iterative normality estimation contribute up to $2.7$ and $1.9$ average points and yield their largest gains on Amazon. The anchor-based reliability gate improves Weibo by $15.6$/$14.9$ points over global Spearman weighting, while the reliability floor adds $2.7$/$2.6$ points on the same target. Together, the components provide complementary gains across target graphs and support the full RINSE configuration.

\paragraph{Analysis of Reliability Gating. }
The gate estimates each view's reliability from its pseudo-AUC on the anchor's ranking extremes (Section~\ref{sec:gate}). Replacing the anchored gate with global Spearman weighting reduces Weibo AUPRC by $15.6$/$14.9$ points under the standard/leak-free protocols, while removing the reliability floor reduces it by $2.7$/$2.6$ points (Figure~\ref{fig:ablation}). These results show that extreme-rank agreement can preserve aligned views while suppressing less reliable evidence.

\paragraph{Sensitivity Analysis. }
Figure~\ref{fig:sensitivity} varies six inference settings around their fixed defaults. In these one-at-a-time sweeps, mean AUPRC across the eight targets changes by less than one point for $\rho\in[0.4,0.8]$, $k\in[5,50]$, and $\tau\in[0,0.2]$. This supports using one fixed configuration across all targets without per-target tuning.

Increasing the estimator rounds from $T=0$ to the performance plateau adds $2.1$ AUPRC points; increasing the ensemble from $E=1$ to $E=5$ adds $3.5$ points; and using two affinity-refinement rounds adds $2.9$ points over no refinement. Further increases provide smaller gains. In particular, $E=8$ adds $0.4$ point over $E=5$, supporting $E=5$ as the default compute-performance trade-off.

\section{Conclusion}

We presented RINSE, a robust, gradient-free target-time framework for zero-shot graph anomaly detection. Around a frozen source-trained scorer, RINSE estimates target normality through trimmed iteration, calibrates embedding statistics, weights complementary evidence by label-free reliability, and rank-averages independently trained encoders. Across eight unseen target graphs, RINSE achieves the highest mean average precision among the evaluated methods under both preprocessing protocols, while the ablation and sensitivity analyses support its individual components. The present study focuses on attributed node-level graphs and a truncated-attention scorer; extending the framework to larger graphs, alternative frozen detectors, and other anomaly settings is a natural direction. Overall, the results support robust target-time estimation as a practical approach to generalist graph anomaly detection.
\section*{Ethical Statement}

RINSE is evaluated on established benchmark datasets and does not collect new human-subject data. It may support the review of potentially fraudulent, spam-related, or manipulative activity without requiring target-domain labels. However, anomaly rankings can contain false positives and may affect users or groups unevenly. They should therefore assist human review rather than trigger automatic punitive decisions. Practical deployment should include domain-specific validation, appropriate threshold selection, privacy safeguards, and subgroup auditing where relevant. RINSE reduces dependence on labeled target data but provides no causal guarantees. Its use should follow applicable laws, platform policies, and data-governance requirements.

\bibliography{main}
\end{document}